%% file: main.tex
\documentclass[journal]{IEEEtran} 

\usepackage[utf8]{inputenc}         
\usepackage[T1]{fontenc}            
\usepackage{hyperref}               
\usepackage{url}                    
\usepackage{nicefrac}               
\usepackage{microtype}              
\usepackage{amsfonts}               
\usepackage{amsmath}                
\usepackage[dvipsnames]{xcolor}     
\usepackage{float}                  
\usepackage{graphicx}               
\graphicspath{{images/}}
\DeclareGraphicsExtensions{.pdf,.PDF,.jpg,.JPG,.jpeg,.JPEG,.png,.PNG}
\usepackage[caption=false]{subfig}  
\usepackage{multicol}               
\usepackage{multirow}               
\usepackage{booktabs}               
\usepackage{array}                  
\newcolumntype{C}[1]{>{\centering\arraybackslash}p{#1}}

\usepackage[nolist]{acronym}        
\usepackage{verbatim}               
\usepackage{comment}                

\usepackage{xcolor}

\begin{document}

\title
{
    Improving Image Autoencoder Embeddings with Perceptual Loss
}

\author{
    \IEEEauthorblockN{
        \textbf{Gustav Grund Pihlgren}, \and 
        \textbf{Fredrik Sandin}, \and 
        \textbf{Marcus Liwicki} \and 
    }\\
    \vspace{0.2cm}
    \IEEEauthorblockA{
        \textit{Lule{\aa} University of Technology, Sweden}\\
        firstname.lastname@ltu.se\\
    }
}


\maketitle

\thispagestyle{empty}

\input{acronym.tex}

\input{sections/0_abstract.tex}

\input{sections/1_introduction.tex}

\input{sections/2_related_work.tex}

\input{sections/3_experimental_setup.tex}

\input{sections/4_results.tex}

\input{sections/5_analysis.tex}

\input{sections/6_conclusion.tex}


\bibliographystyle{IEEEtran}
\bibliography{biblio}

\end{document}

%% file: acronym.tex
\begin{acronym}[Bash]
    
    \acro{AI}{Artificial Intelligence}
    \acro{ANN}{Artificial Neural Network}
    
    \acro{CNN}{Convolutional Neural Network}
    
    \acro{MLP}{Multilayer Perceptron}
    
    \acro{NC-Net}{Neighbourhood Consensus Network}
    
    \acro{RL}{Reinforcement Learning}
    
\end{acronym}

%% file: sections/0_abstract.tex
\begin{abstract}


%
%
%
%
%

Autoencoders are commonly trained using element-wise loss.
However, element-wise loss disregards high-level structures in the image which can lead to embeddings that disregard them as well.
A recent improvement to autoencoders that helps alleviate this problem is the use of perceptual loss.
This work investigates perceptual loss from the perspective of encoder embeddings themselves.
Autoencoders are trained to embed images from three different computer vision datasets using perceptual loss based on a pretrained model as well as pixel-wise loss.
A host of different predictors are trained to perform object positioning and classification on the datasets given the embedded images as input.
The two kinds of losses are evaluated by comparing how the predictors performed with embeddings from the differently trained autoencoders.
The results show that, in the image domain, the embeddings generated by autoencoders trained with perceptual loss enable more accurate predictions than those trained with element-wise loss.
Furthermore, the results show that, on the task of object positioning of a small-scale feature, perceptual loss can improve the results by a factor 10.
The experimental setup is available online. \footnote{\url{https://github.com/guspih/Perceptual-Autoencoders}}



\end{abstract}

%% file: sections/1_introduction.tex
\section{Introduction}
\label{toc:introduction}




Autoencoders have been in use for decades~\cite{rumelhart1985learning, ballard1987modular} and are prominently used in machine learning research today~\cite{NIPS2018_7512, Kusner:2017:GVA:3305381.3305582, gosztolya2019autoencoderbased}.
Autoencoders have been commonly used for feature learning and dimensionality reduction~\cite{Goodfellow-et-al-2016}.
The reduced dimensions are referred to as the latent space, embedding, or simply as $z$.
However, autoencoders have also been used for a host of other tasks like generative modeling~\cite{yuhuai2016generative}, denoising~\cite{vincent2008denoising}, generating sparse representations~\cite{ranzato2007sparse}, anomaly detection~\cite{hawkins2002outlier} and more.

Traditionally, autoencoders are trained by making the output similar to the input. 
The difference between output and input is quantified by a loss function, like Mean Squared Error (MSE), applied to the differences between each output unit and their respective target output.
This kind of loss, where the goal of each output unit is to reconstruct exactly the corresponding target, is known as element-wise loss.
In computer vision, when the targets are pixel values, this is known as pixel-wise loss, a specific form of element-wise loss.

\begin{figure}[t] 
    \centering
    \includegraphics[width=0.6\columnwidth]{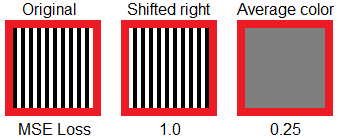}
    \caption{A striped image and two reconstructions with their respective element-wise Mean Squared Error. In the first reconstruction each stripe has been moved one pixel to the side and the other is completely gray.}
    \label{fig:pixel_vs_perceptual}
\end{figure}

A problem related to element-wise loss is that it does not take into account relations between the different elements; it only matters that each output unit is as close as possible to the corresponding target.
This problem is visualized in Fig.~\ref{fig:pixel_vs_perceptual}, where the first reconstruction with loss $1.0$ is the correct image shifted horizontally by one pixel, and the second reconstruction with lower loss has only one color given by the mean value of the pixels.
While a human would likely say that the first reconstruction is more accurate, pixel-wise loss favors the latter.
This is because, for a human, the pattern is likely more important than the values of individual pixels.
Pixel-wise loss does, on the other hand, only account for the correctness of individual pixels.

\begin{figure}[t] 
    \centering
    \includegraphics[width=0.8\columnwidth]{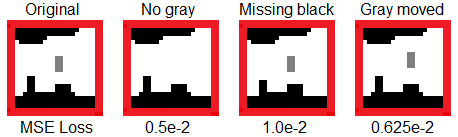}
    \caption{An image and three reconstructions with their respective element-wise Mean Squared Error. The first is the original image. The second is missing the gray feature. The third has four black pixels removed. In the last the gray feature have been moved one pixel up and one pixel to the right.}
    \label{fig:important_feature}
\end{figure}

Another problem with element-wise loss is that all elements are weighted equally although some group of elements may be more important,
for example when solving computer vision tasks like object detection.
This problem is visualized in Fig.~\ref{fig:important_feature} where an otherwise black and white image has a small gray feature.
Despite being perceived as important by humans, element-wise loss gives only a small error for completely omitting the gray feature.
This is because element-wise loss considers each element to have the same importance in reconstruction, even though some elements might represent a significant part of the input space.

One method that has been used to alleviate these problems for image reconstruction and generation is perceptual loss.
Perceptual loss is calculated using the activations of an independent neural network, called the perceptual loss network.
This type of loss was introduced for autoencoders in~\cite{pmlr-v48-larsen16}.

Despite the success of perceptual loss for autoencoders when it comes to image reconstruction and generation, the method has yet to be tested for its usefulness for maintaining information in embedded data in the encoding-decoding task itself.
This work investigates how training autoencoders with perceptual loss affects the usefulness of the embeddings for the tasks of image object positioning and classification.
This is done by comparing autoencoders and variational autoencoders (VAE) trained with pixel-wise loss to those trained with perceptual loss.

\subsection*{Contribution}

This work shows that, on three different datasets, embeddings created by autoencoders and VAEs trained with perceptual loss are more useful for solving object positioning and image classification than those created by the same models trained with pixel-wise loss.
This work also shows that if an image has a small but important feature, the ability to reconstruct this feature from embeddings can be greatly improved if the encoder has been trained with perceptual loss compared to pixel-wise loss.

%% file: sections/2_related_work.tex
\section{Related Work}
\label{toc:related_Work}

The VAE is an autoencoder architecture that has seen much use recently~\cite{Kingma2013AutoEncodingVB}.
The encoder of a VAE generates a mean and variance of a Gaussian distribution per dimension instead of an embedding.
The embedding is then sampled from those distributions.
To prevent the model from setting the variance to $0$ a regulatory loss based on Kullback-Liebler (KL) divergence~\cite{kullback1951information} is used.
This regulatory loss is calculated as the KL divergence between the generated Gaussian distributions and a Gaussian with mean $0$ and variance $1$.
Through this the VAE is incentivized not only to create embeddings that contain information about the data, but that these embeddings closely resemble Gaussians with mean $0$ and variance $1$.
The balance between the reconstruction and KL losses incentivizes the VAE to place similar data close in the latent space which means that if you sample a point in the latent space close to a know point those are likely to be decoded similarly.
This sampling quality of the VAE makes it a good generative model in addition to its use as feature learner.

The VAE has been combined with another generative model, the GAN~\cite{goodfellow2014generative} to create the VAE-GAN~\cite{pmlr-v48-larsen16}.
In order to overcome problems with the VAE as generative model the VAE-GAN adds a discriminator to the architecture, which is trained to determine if an image have been generated or comes from the ground truth.
The VAE is then given an additional loss for fooling the discriminator and a perceptual loss by comparing the activations of the discriminator when given the ground truth to when it is given the reconstruction.
This means that the discriminator network is also used as a perceptual loss network in the VAE-GAN.
While the VAE-GAN was the first autoencoder to use perceptual loss, it was not the first use of perceptual loss.

Perceptual loss was introduced by the field of explainable AI as a way to visualize the optimal inputs for specific classes or feature detectors in a neural network~\cite{simonyan2013deep, DBLP:journals/corr/YosinskiCNFL15}.
Soon after, GAN were introduced which used perceptual loss to train a generator network to fool a discriminator network.
In order to use perceptual loss without the need for training a discriminator, which can be notoriously difficult, \cite{dosovitskiy2016generating} proposed using image classification networks in their place.
In that work AlexNet~\cite{alexnet2012} is used as perceptual loss network.

While the use of perceptual loss has been primarily to improve image generation it has also been used for image segmentation~\cite{mosinska2017beyond}, object detection~\cite{Li_2017_CVPR}, and super-resolution~\cite{johnson2016style}.

%% file: sections/3_experimental_setup.tex
\section{Perceptual Loss}
\label{toc:perceptual_loss}

Perceptual loss is in essence any loss that depends on some activations of a neural network beside the machine learning model that is being trained.
This external neural network is referred to as the perceptual loss network.
In this work perceptual loss is used to optimize the similarity between the image and its reconstruction as perceived by the perceptual loss network.
By comparing feature extractions of the perceptual loss network when it's given the original input compared to the recreation, a measure for the perceived similarity is created.
This process is described in detail below.

Given an input $X$ of size $n$ an autoencoder can be defined as a function $\hat{X} = a(X)$ where $\hat{X}$ is the reconstruction of $X$.
Given a loss function $f$ (like square error or cross entropy) the element-wise loss for $a$ is defined as:
\begin{equation}
    E = \sum^n_{k=1} f(X_k, a(X)_k)
\end{equation}

Given a perceptual loss network $y = p(X)$ where $y$ is the relevant features of size $m$ the perceptual loss for $a$ is defined as:
\begin{equation}
    E = \sum^m_{k=1} f(p(X)_k, p(a(X))_k)
\end{equation}

Optionally the average can be used instead of the sum.

\begin{figure}[ht] 
    \includegraphics[width=\columnwidth]{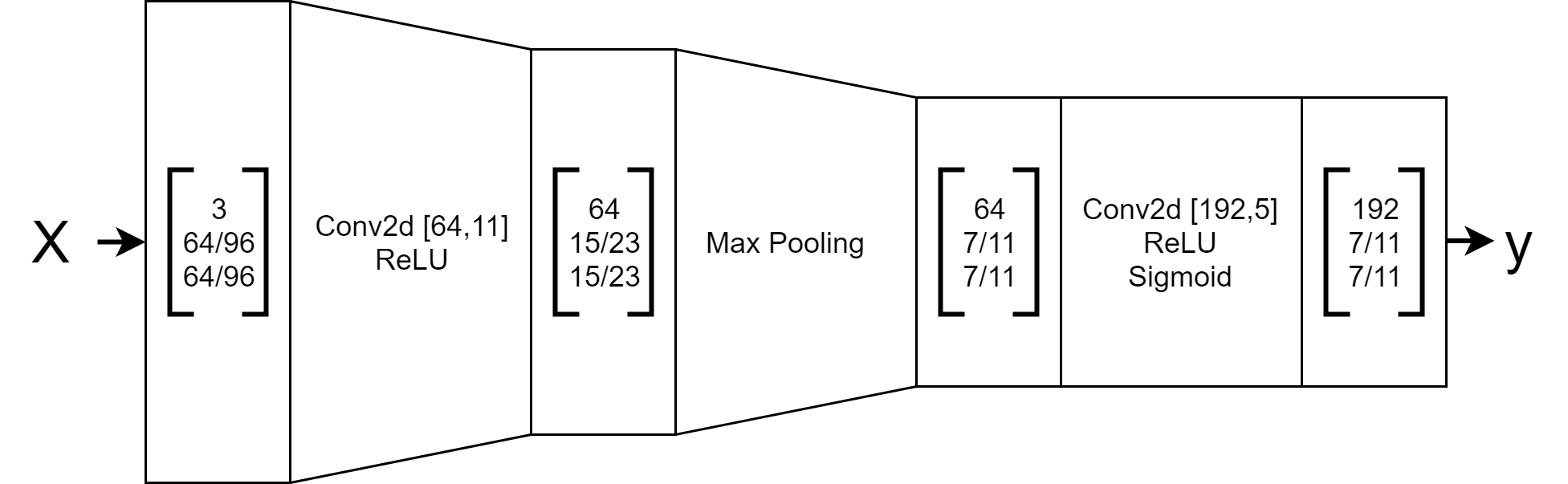}
    \caption{The parts of a pretrained AlexNet that were used for calculating and backpropagating the perceptual loss.}
    \label{fig:image_alexnet_loss}
\end{figure}

This work, like a previous work~\cite{dosovitskiy2016generating}, uses AlexNet~\cite{krizhevsky2014one} pretrained on ImageNet~\cite{deng2009imagenet} as perceptual loss network ($p$).
For this work, feature extraction is done early in the convolutional part of the network since we are interested in retaining positional information which would be lost by passing through too many pooling or fully-connected layers.
With that in mind feature extraction of $y$ from AlexNet is done after the second ReLU layer.
To normalize the output of the perceptual loss network a sigmoid function was added to the end.
The parts of the perceptual loss network that are used are visualized in Fig.~\ref{fig:image_alexnet_loss}.

\section{Datasets}
\label{toc:datasets}

This work makes use of three image datasets each with a task that is either object positioning or classification. These three are a collection of images from the LunarLander-v2 environment of OpenAI Gym~\cite{brockman2016openai}, \textit{STL-10}~\cite{coates2011analysis}, and \textit{SVHN}~\cite{netzer2011reading}.

\subsection{LunarLander-v2 collection}

The \textit{LunarLander-v2 collection} consists of images collected from $1400$ rollouts of the LunarLander-v2 environment using a random policy.
Each rollout is $150$ timesteps long.
The images are scaled down to the size of $64\times64$ pixels.
The first $700$ rollouts are unaltered while all images where the lander is outside the screen have been removed from the remaining rollouts.
This process removed roughly $10\%$ of the images in the latter half.
The task of the \textit{LunarLander-v2 collection} is object positioning, specifically to predict the coordinates of the lander in the image.

\subsection{STL-10}
The \textit{STL-10} dataset consists of $100000$ unlabeled images, $500$ labeled images, and $8000$ test images of animals and vehicles.
The labeled and test images are divided into $10$ classes.
The task is to classify the images.
Specifically the task is to create a model that uses unsupervised learning on the unlabelled images to complement training on the few labelled samples.
The images are of size $96\times96$.

\textit{NOTE:} The original task of STL-10 is to only use the provided training data to train a classifier.
However, the AlexNet part of the perceptual loss that this work uses has been pretrained on another dataset.
Thus, any results achieved by a network trained with that loss cannot be regarded as actually performing the original task of STL-10.

\subsection{SVHN}
The \textit{SVHN} dataset consists of images of house numbers where the individual digits have been cropped out and scaled to $32\times32$ pixels.
The dataset consists of $73257$ training images, $26032$ testing images, and $531131$ extra images.
The task is to train a classifier for the digits.
The extra images are intended as additional training data if needed.

\section{Autoencoder Architecture}
\label{toc:autoencoder_architecture}

\begin{figure}[ht] 
    \includegraphics[width=\columnwidth]{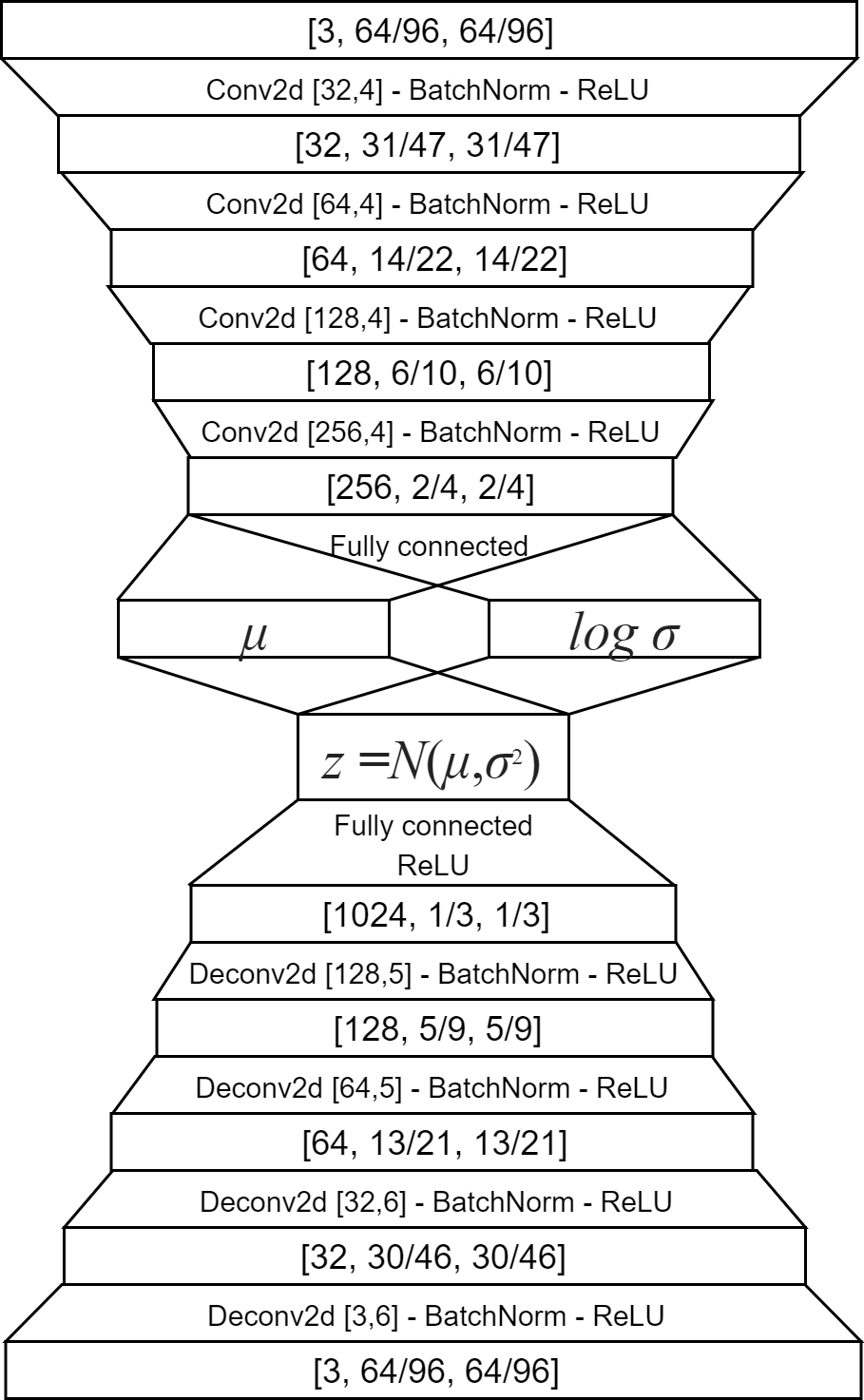}
    \caption{The convolutional variational autoencoder used in this work.}
    \label{fig:architecture}
\end{figure}

The autoencoder architecture used in this paper is the same for all datasets and is based on the architecture in~\cite{NIPS2018_7512} and the full architecture can be seen in Fig.~\ref{fig:architecture}.
The architecture takes input images of size $3\times64\times64$ or $3\times96\times96$.
For the SVHN dataset the input to the architecture is each images duplicated into a 2-by-2 grid of $64\times64$ pixels.
The stride for all convolutional and deconvolutional layers is $2$.
When training a standard (non-variational) autoencoder $\sigma$ and the KLD-loss is set to $0$.

\section{Testing Procedure}
\label{toc:testing_procedure}
For each dataset a number of autoencoders were trained with different numbers of latent dimensions.
Since the use of autoencoder embeddings to minimize the input for a task is typically helpful when data or labels are limited this work investigates small sizes of the latent space ($\sim100$).
With the actual sizes ($z$) tested being $32$, $64$, $128$, $256$, $512$.
Not all values of $z$ were tested for all datasets, with smaller values used for datasets with lower dimensionality.

For each size of the latent space two standard autoencoders and two VAEs were trained.
One of each with pixel-wise loss (AE and VAE) and one of each with perceptual loss (P. AE and P. VAE).
Then for each trained autoencoder a number of predictors were trained to solve the task of that dataset given the embedding as input.
There were two types of predictors; (i) Multilayer Perceptrons (MLP) with varying parameters and (ii) linear regressors.

\begin{figure}[ht] 
    \includegraphics[width=\columnwidth]{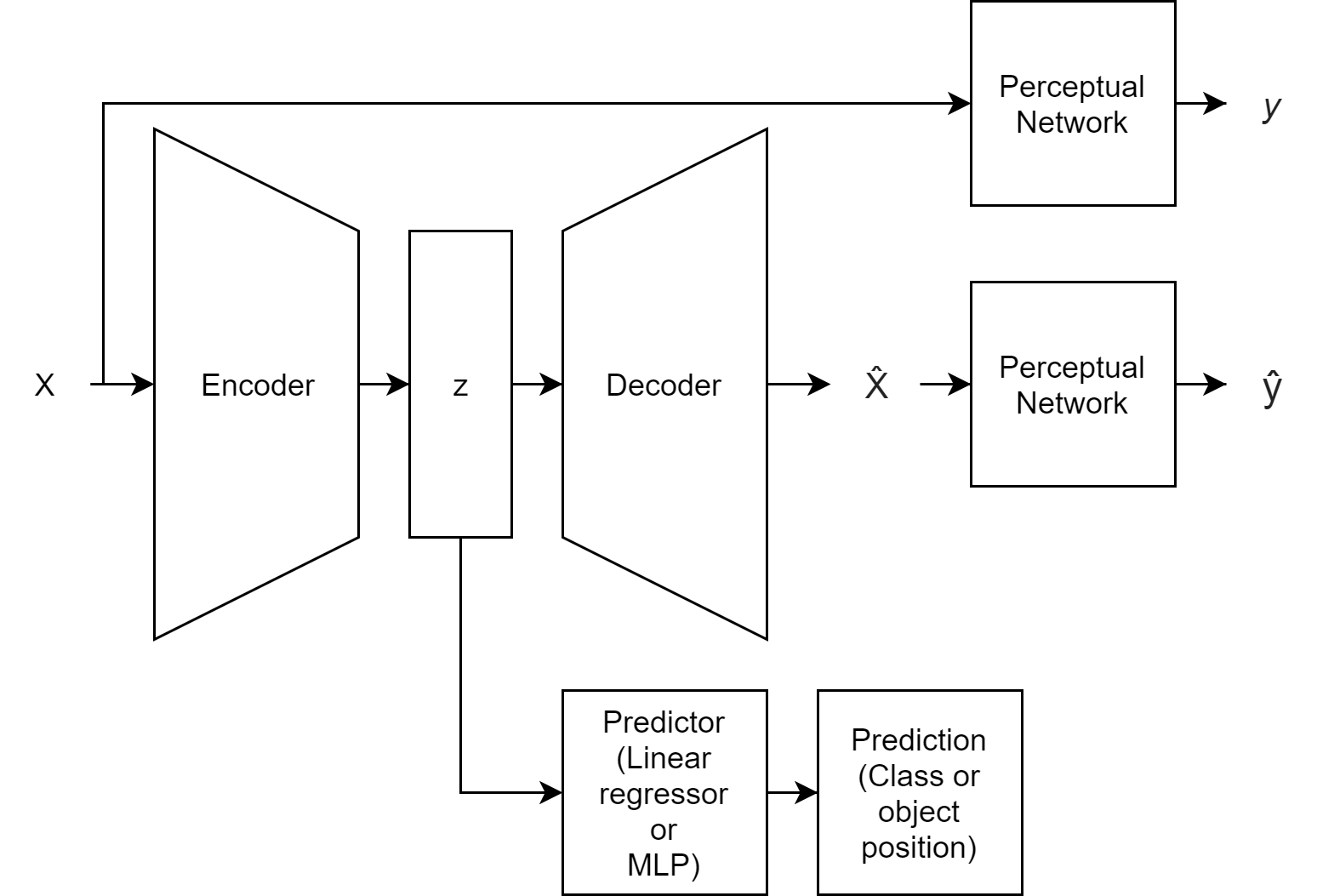}
    \caption{The system used in this work including both the autoencoding and prediction pathways.}
    \label{fig:system}
\end{figure}

The full system including the predictor is shown in Fig.~\ref{fig:system}.
The encoder, $z$, and decoder make up the autoencoder which is shown in Fig.~\ref{fig:architecture}.
The autoencoder is either trained with pixel-wise loss given by MSE between $X$ and $\hat{X}$, or perceptual loss given by MSE between $y$ and $\hat{y}$.
The perceptual loss network is the part of AlexNet that is detailed in Fig.~\ref{fig:image_alexnet_loss}.
The predictor is either a linear regressor or an MLP as described below.

The MLPs had 1 or 2 hidden layers with 32, 64, or 128 hidden units each and with ReLU or Sigmoid activation functions.
The output layer either lacked activation function or used Softmax.
The entire search space of hyperparameters were considered with the restriction that the second hidden layer couldn't be larger than the first.

Each dataset is divided into three parts:
One for training and validating the autoencoders, one for training and validating the predictors, and one for testing.
Table~\ref{tab:data_disribution} shows which parts of each dataset are used for what part of the evaluation.
Of the autoencoder and predictor parts 80\% is used for training and 20\% is used for validation.

\begin{table}[ht]
    \caption{Which parts of the datasets are used for training the autoencoders and predictors, and which is used for final testing.}
    \label{tab:data_disribution}
    \begin{center}
    \begin{tabular}{l c c c}
    \toprule
                        & LunarLander   & STL-10    & SVHN      \\\hline
        Autoencoder     & Unaltered     & Unlabeled & Extra     \\
        Predictor       & 80\% altered  & Training  & Training  \\
        Test            & 20\% altered  & Test      & Test      \\
    \bottomrule
    \end{tabular}
    \end{center}
\end{table}

For each trained autoencoder the MLP and regressor with the lowest validation loss was tested using the test set.
For the LunarLander-v2 collection the results are the distance between the predicted position and the actual position averaged over the test set and for the other datasets the results are the accuracy of the predictor on classifying the test set.
Additionally the decoders of all autoencoders were retrained with pixel-wise MSE loss to see if the reconstructions using perceptually trained embeddings would be better than with pixel-wise trained embeddings.

%% file: sections/4_results.tex
\section{Results}
\label{toc:results}

The results are broken into seven tables.
Tables~\ref{tab:lunar_results},~\ref{tab:stl10_results}, and~\ref{tab:svhn_results} show the performance of the MLPs with the lowest validation loss on the test sets.
Tables~\ref{tab:lunar_regressor_results},~\ref{tab:stl10_regressor_results}, and~\ref{tab:svhn_regressor_results} show the performance of the regressors on the test sets.
For reference the state-of-the-art accuracy, at the time of writing, on STL-10 and SVHN are $94.4\%$~\cite{berthelot2019mixmatch} and $99.0\%$~\cite{cubuk2018autoaugment} respectively.
Table~\ref{tab:reconstruction_score} shows the relative performance on image reconstruction (as measured with the L1 norm) for the best of each type of autoencoder on each dataset.

\begin{figure}[ht] 
    \includegraphics[width=\columnwidth]{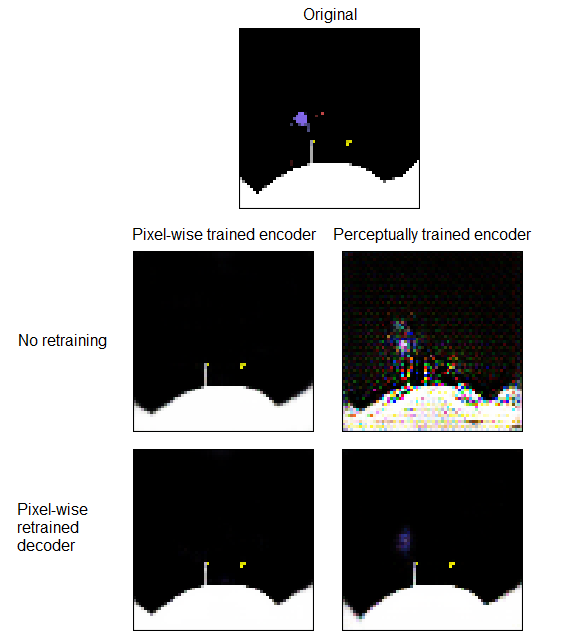}
    \caption{A comparison of the reconstructed images from pixel-wise and perceptually trained embeddings.}
    \label{fig:retrained_decoder}
\end{figure}

Actual reconstructed images from the LunarLander-v2 collection are visualized in Fig.~\ref{fig:retrained_decoder}.
The image contains the original image and its reconstructions by a pixel-wise and a perceptually trained autoencoder before and after retraining.
For this image the reconstruction by the perceptual autoencoder has higher pixel-wise reconstruction error before as well as after retraining of the decoder.

Over all tests, the use of perceptual loss added $12\pm3\%$ to the training time of the autoencoders. Since the perceptual loss is only used during autoencoder training it had no effect on the time for inference or training predictors.

\begin{table}[ht]
    \caption{Average test distance error in pixels for LunarLander-v2 collection for the MLPs with lowest validation loss.}
    \label{tab:lunar_results}
    \begin{center}
    \begin{tabular}{l c c c c}
    \toprule
        z size  & AE    & VAE   & P. AE & P. VAE        \\\hline
        32      & 13.44 & 12.70 & 2.11  & \textbf{1.28} \\
        64      & 13.28 & 13.22 & 1.60  & \textbf{1.15} \\
        128     & 13.31 & 13.31 & 1.44  & \textbf{1.15} \\
        256     & 13.22 & 13.28 & 1.44  & \textbf{1.28} \\
        Any     & 13.22 & 12.70 & 1.44  & \textbf{1.15} \\
    \bottomrule
    \end{tabular}
    \end{center}
\end{table}

\begin{table}[ht]
    \caption{Accuracy on STL-10 test set for various z sizes for the MLPs with lowest validation loss.}
    \label{tab:stl10_results}
    \begin{center}
    \begin{tabular}{l c c c c}
    \toprule
        z size  & AE        & VAE       & P. AE     & P. VAE            \\\hline
        64      & 38.5\%    & 41.9\%    & 61.4\%    & \textbf{63.4\%}   \\
        128     & 39.0\%    & 42.4\%    & \textbf{62.6\%}   & 62.3\%    \\
        256     & 41.5\%    & 38.9\%    & 63.4\%    & \textbf{64.8\%}   \\
        512     & 45.0\%    & 39.2\%    & 63.3\%    & \textbf{64.4\% }  \\
        Any     & 45.0\%    & 42.4\%    & 63.4\%    & \textbf{64.8\%}   \\
    \bottomrule
    \end{tabular}
    \end{center}
\end{table}

\begin{table}[ht]
    \caption{Accuracy on SVHN test set for various z sizes for the MLPs with lowest validation loss.}
    \label{tab:svhn_results}
    \begin{center}
    \begin{tabular}{l c c c c}
    \toprule
        z size  & AE        & VAE       & P. AE             & P. VAE    \\\hline
        32      & 76.5\%    & 76.8\%    & \textbf{77.0\%}   & 69.6\%    \\
        64      & 81.5\%    & 81.8\%    & \textbf{82.2\% }  & 76.8\%    \\
        128     & 81.9\%    & 82.7\%    & \textbf{84.0\%}   & 81.2\%    \\
        Any     & 81.9\%    & 82.7\%    & \textbf{84.0\%}   & 81.2\%    \\
    \bottomrule
    \end{tabular}
    \end{center}
\end{table}

\begin{table}[ht]
    \caption{Average test distance error in pixels for LunarLander-v2 collection for the regressors with lowest validation loss.}
    \label{tab:lunar_regressor_results}
    \begin{center}
    \begin{tabular}{l c c c c}
    \toprule
        z size  & AE    & VAE   & P. AE & P. VAE        \\\hline
        32      & 13.73 & 12.67 & \textbf{7.23} & 7.46  \\
        64      & 13.60 & 12.99 & 6.85  & \textbf{5.50} \\
        128     & 14.18 & 12.96 & 6.72  & \textbf{5.54} \\
        256     & 15.17 & 13.38 & 6.40  & \textbf{5.18} \\
        Any     & 13.60 & 12.67 & 6.40  & \textbf{5.18} \\
    \bottomrule
    \end{tabular}
    \end{center}
\end{table}

\begin{table}[ht]
    \caption{Accuracy on STL-10 test set for various z sizes for the regressors with lowest validation loss.}
    \label{tab:stl10_regressor_results}
    \begin{center}
    \begin{tabular}{l c c c c}
    \toprule
        z size  & AE        & VAE       & P. AE     & P. VAE            \\\hline
        64      & 34.6\%    & 36.8\%    & 56.3\%    & \textbf{58.1\%}   \\
        128     & 35.4\%    & 39.7\%    & \textbf{59.7\%}    & 57.4\%   \\
        256     & 40.2\%    & 43.6\%    & 59.5\%    & \textbf{63.7\%}   \\
        512     & 44.4\%    & 45.9\%    & 60.1\%    & \textbf{64.9\%}   \\
        Any     & 44.4\%    & 45.9\%    & 60.1\%    & \textbf{64.9\%}   \\
    \bottomrule
    \end{tabular}
    \end{center}
\end{table}

\begin{table}[h!]
    \caption{Accuracy on SVHN test set for various z sizes for the regressors with lowest validation loss.}
    \label{tab:svhn_regressor_results}
    \begin{center}
    \begin{tabular}{l c c c c}
    \toprule
        z size  & AE        & VAE       & P. AE     & P. VAE            \\\hline
        32      & 25.5\%    & 32.0\%    & \textbf{61.6\%}    & 58.9\%   \\
        64      & 24.8\%    & 41.2\%    & \textbf{67.2\% }   & 66.3\%   \\
        128     & 29.3\%    & 47.5\%    & 70.6\%    & \textbf{71.2}\%   \\
        Any     & 29.3\%    & 47.5\%    & 70.6\%    & \textbf{71.2}\%   \\
    \bottomrule
    \end{tabular}
    \end{center}
\end{table}

\begin{table}[h!]
    \caption{Performance of reconstruction of the best autoencoders after retraining as measured by the performance relative to the autoencoder with the lowest L1-norm error.}
    \label{tab:reconstruction_score}
    \begin{center}
    \begin{tabular}{l c c c c}
    \toprule
        Dataset     & AE    & VAE   & P. AE & P. VAE    \\\hline
        LunarLander & 93\%  & 77\%  & 100\% & 73\%      \\
        STL-10      & 100\% & 92\%  & 60\%  & 44\%      \\
        SVHN        & 100\% & 97\%  & 75\%  & 57\%      \\
    \bottomrule
    \end{tabular}
    \end{center}
\end{table}

%% file: sections/5_analysis.tex
\section{Analysis}
\label{toc:analysis}

The most prominent result of the experiments is that for all three datasets and all tested sizes of $z$ the perceptually trained autoencoders performed better than the pixel-wise trained ones.
Furthermore, for both the LunarLander-v2 collection and STL-10 the pixel-wise trained autoencoder is outperformed significantly.
On the LunarLander-v2 collection the perceptually trained autoencoders have an order of magnitude better performance.

While the performance on object positioning and classification is better for perceptually trained autoencoders this is not the case with image reconstruction.
On LunarLander-v2 collection the perceptual autoencoder is only slightly better at reconstruction.
For the other two datasets the pixel-wise trained autoencoders have a much lower relative reconstruction error.
Furthermore in Fig.~\ref{fig:retrained_decoder} the reconstructed image where the lander is actually visible has higher reconstruction loss.
This is an actual demonstration of the problems with pixel-wise reconstruction metrics that were visualized in Fig~\ref{fig:pixel_vs_perceptual} and Fig.~\ref{fig:important_feature}.
This lack of correlation between low reconstruction error and performance on a given task is in line with the findings of~\cite{alberti2017pitfall}.

The results suggest that perceptual loss gives, for the tasks at hand, better embeddings than pixel-wise loss.
Taking it even further however, these results combined with earlier work~\cite{pmlr-v48-larsen16} suggests that pixel-wise reconstruction error is a flawed way of measuring the similarity of two images.

However, while the results are better for perceptual loss this comes at the cost of training time.
While a 12\% increase of training time is not a significant amount, especially since training and inference of downstream tasks is not noticeably affected, this increase depends on the perceptual loss network's size in comparison to the size of the remaining model.
If the autoencoder is small or the perceptual loss network significantly large the effect on training time could become significant.

Another interesting aspect is the difference in performance when switching from MLPs to linear regression.
The error of perceptually trained autoencoders on LunarLander-v2 collection is increased by a factor 5 when switching from MLPs to linear regression.
This suggests that while the embeddings of the perceptually trained autoencoders contains much more details as to the location of the lander, this information is not encoded linearly which makes linear regressors unable to extract it properly.
This is in contrast to STL-10 on which the performance remains roughly the same for both predictor types, which suggests that all the information needed to make class prediction is encoded linearly. 

On SVHN performance were similar for all autoencoders with MLP predictiors.
However, the performance of pixel-wise trained AE and VAE lose 50 and 40 percentage points respectively when using linear regression.
This indicates that the autoencoders manage to encode similarly useful information for solving the task but that the pixel-wise trained embeddings demand a non-linear model to extract that information.

All this comes together to suggests that perceptually trained autoencoders either have more useful embeddings or the useful information in the embeddings require less computational resources to make use of.
The accessibility of the information is important as one of the primary uses of autoencoders is dimensionality reduction to enable the training of smaller predictors for the task at hand.
If the information is computationally heavy to access a significant part of the already small model may be dedicated to unpacking it instead of doing prediction.

It is important to note the scope of this work.
Only three datasets have been investigated, and for only a single perceptual loss network.
The work shows that there is promise in investigating this use of perceptual loss, but further studies are needed.
Specifically to see if these results generalize to other datasets and perceptual loss networks.


%% file: sections/6_conclusion.tex
\section{Conclusion}
\label{toc:conclusion}


Element-wise loss disregards high-level features in images which can lead to embeddings that do not encode the features sufficiently well.
This work investigates perceptual loss as an alternative to element-wise loss to improve autoencoder embeddings for downstream prediction tasks.
The results show that perceptual loss based on a pretrained model produces better embeddings than pixel-wise loss for the three tasks investigated.
This work demonstrates that it is important to research on alternatives to element-wise loss and to directly analyze the learned embeddings.
Future investigations of perceptual loss should investigate the importance of which perceptual loss network one chooses, how the features are extracted, and in general apply perceptual loss in other domains than images. 